\documentclass[conference]{IEEEtran}
\usepackage{cite}
\usepackage{amsmath,amssymb,amsfonts}
\usepackage{algorithmic}
\usepackage{graphicx}
\usepackage{textcomp}
\usepackage{xcolor}
\usepackage{capt-of}
\usepackage[switch]{lineno} 
\usepackage{array}
\newcommand{\PreserveBackslash}[1]{\let\temp=\\#1\let\\=\temp}
\newcolumntype{C}[1]{>{\PreserveBackslash\centering}p{#1}}
\newcolumntype{R}[1]{>{\PreserveBackslash\raggedleft}p{#1}}
\newcolumntype{L}[1]{>{\PreserveBackslash\raggedright}p{#1}}

\def\BibTeX{{\rm B\kern-.05em{\sc i\kern-.025em b}\kern-.08em
    T\kern-.1667em\lower.7ex\hbox{E}\kern-.125emX}}
\begin{document}

\title{An AI Architecture with the Capability to Explain Recognition Results}

\author{\IEEEauthorblockN{Paul Whitten, Francis Wolff, Chris Papachristou}
\IEEEauthorblockA{\textit{Electrical, Computer, and Systems Engineering} \\
\textit{Case School of Engineering}\\
\textit{Case Western Reserve University}\\
Cleveland, OH, USA\\
pcw@case.edu, fxw12@case.edu, cap2@case.edu}

}

\maketitle

\begin{abstract}

Explainability is needed to establish confidence in machine learning results.
Some explainable methods take a post hoc approach to explain the weights of
machine learning models, others highlight areas of the input contributing to
decisions. These methods do not adequately explain decisions, in plain terms.
Explainable property-based systems have been shown to provide explanations in
plain terms, however, they have not performed as well as leading unexplainable
machine learning methods. This research focuses on the importance of metrics to
explainability and contributes two methods yielding performance gains. The first
method introduces a combination of explainable and unexplainable flows,
proposing a metric to characterize explainability of a decision.  The second
method compares classic metrics for estimating the effectiveness of neural
networks in the system, posing a new metric as the leading performer.  Results
from the new methods and examples from handwritten datasets are presented.

\end{abstract}


\section{Introduction}

Explainability is needed to establish confidence in machine learning (ML) results.
Much work has been posed to assist in explaining automated decisions by
Artificial Intelligence (AI), however, current explainable results still remain
unsatisfactory.

Post hoc analysis of Neural Networks (NN) has assessed local and global
decisions by examining the weights of the NN. Explainability through the
identification of areas of the input that contribute to a decision has been
posed.  While these add an element of explainability, they have been unable to
explain, in plain terms, why a decision was made by an automated system.




The focus of this research is the use of techniques to provide explainable
results with combined property-based NN models. Results center on explainability
and are not meant to compete with the latest unexplainable recognition
techniques.

\begin{figure}
    \includegraphics[width=9.0cm]{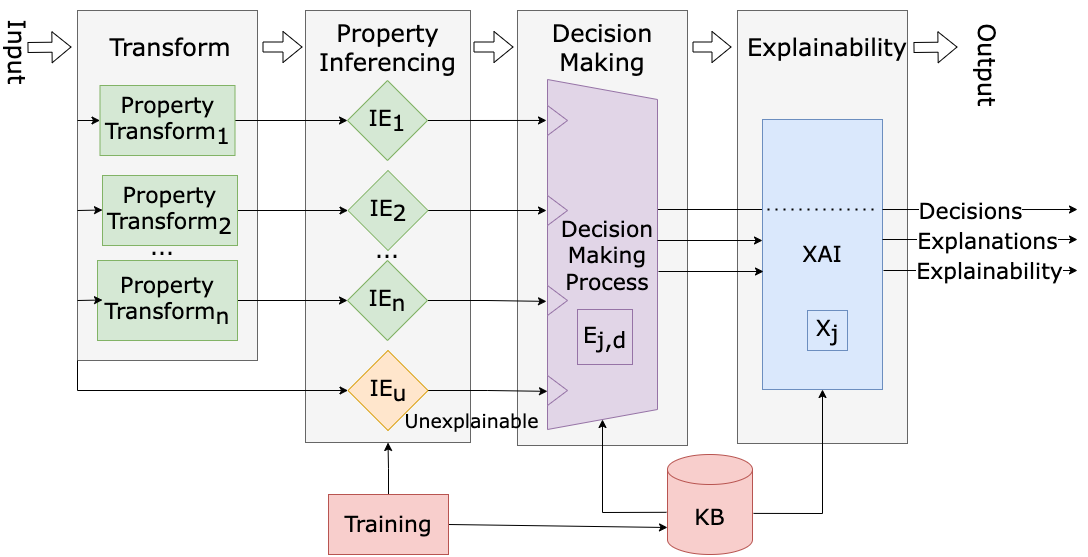}
    \caption{An explainable architecture with an unexplainable flow.}
    \label{fig:xai_arch}
\end{figure}


Contributions of this work include two methods to achieve explainability and
performance gains.  The first is the use of an unexplainable ML model in an
explainable architecture shown in Fig.~\ref{fig:xai_arch}.  Unexplainability
occurs when a property cannot account for an explanation. Unexplainability
becomes an additional component in the architecture. In order to improve the
overall explainability and results, and to quantify explainability of decisions,
a new explainability metric, $Ex_d$, is introduced.  The second is a comparison
of classic ML performance metrics and proposing the $E_{PARS}$ metric to improve
accuracy in an explainable architecture.

The popular MNIST and EMNIST handwritten digit and character datasets
\cite{deng2012mnist, cohen2017emnist} are leveraged to explore the new methods
in recognizing handwritten characters.  With these datasets, the problem
approached is one of recognizing the class, $d$, of an input, where class is
defined as a particular digit in MNIST or an alphanumeric character in
EMNIST. 





Related work is outlined in Section \ref{related_work}.  Section \ref{method}
presents the explainable architecture, the addition of unexplainable flows to
combined NN models to increase performance, and introduces per-class effectiveness
metrics. Finally, results including an explainable example are discussed in
Section \ref{results}.


\section{Related Work} \label{related_work}

Several works discuss the combination of results of multiple trained neural
networks. Jacobs, et al. identified local experts of trained networks
\cite{6797059}. Other works treated multiple trained networks as committees and
combined NNs to form a collective decision \cite{perrone1993putting,
sharkey1996combining}.

An explainable additive NN model is posed by Vaughan et al. where a
system is composed of layering distinct NNs that are trained on transforms of
the inputs.  A layer then combines the outputs of the distinct NNs to perform a
prediction. Explainability comes from each distinct NN modeling features of the
input which lends to interpretability of the architecture
\cite{vaughan2018explainable}.

An explainable architecture using explainable properties, transforms of input
related to the properties, Inference Engines (IE) for each property,
probabilistic decision-making, and attributing decisions to relevant explainable
properties was posed\cite{whitten21, whitten23}.  Like combined NN and additive NN
systems, the explainable architecture examined decisions of multiple NNs.


Metrics such as Accuracy, Precision, and Recall have evolved from disciplines such as
Statistics, Data Science, and Information Retrieval\cite{harter1971cranfield} to
use in the evaluation of ML models.  The Accuracy metric \eqref{eq:accuracy} for
a binary predictor is given by the ratio of correct predictions (positive and
negative), the sum of True Positives ($TP$) and True Negatives ($TN$), to the
total number of predictions which is given by the sum of $TP$, $TN$, False Positives
($FP$), and False Negatives ($FN$).  The Precision metric \eqref{eq:precision}
represents the ratio of correct positive predictions, $TP$, to all positive
predictions, $TP + FP$.  Recall \eqref{eq:recall} represents the ratio of
correct positive predictions, $TP$, to all positive cases, $TP + FN$.
Specificity \eqref{eq:specificity} represents the ratio of correct negative
predictions, $TN$, to all negative cases, $TN + FP$.

There has been extensive study of additional metrics for gauging the performance
of ML models \cite{picek2019curse, erickson2021magician, Naser_2021}.  The
Receiver Operating Characteristic (ROC) Curve provides a relationship between
the $TP$ and $FP$ Rates for various thresholds and the Area Under the ROC Curve
(AUC) \cite{METZ1978283, Hanley_1982} provides a metric characterizing the
performance of a classifier. F1-Score is a metric that uses the harmonic mean of
Recall and Precision of a model \cite{sasaki2007truth}. Cohen's Kappa was used
as a metric for characterizing the agreement between observers of psychological
behavior \cite{Cohen1960ACO,ben2008relationship}. In the case of ML models
Cohen's Kappa is effective in comparing agreement between labels and
predictions. Matthew's correlation coefficient (MCC) is presented as a leading
metric for comparison of imbalanced datasets \cite{MATTHEWS1975442, 9385097}.
AUC, F1-Score, Cohen's Kappa, and MCC are compared as potential effectiveness
metrics in the explainable architecture.

\begin{equation}\label{eq:accuracy}
    Accuracy = ACC = \frac{TP + TN}{TP + TN + FP + FN}
\end{equation}

\begin{equation}\label{eq:precision}
    Precision = P = \frac{TP}{TP + FP}
\end{equation}

\begin{equation}\label{eq:recall}
    Recall = R = \frac{TP}{TP + FN}
\end{equation}

\begin{equation}\label{eq:specificity}
    Specificity = S = \frac{TN}{TN + FP}
\end{equation}

\begin{table}
    \renewcommand{\arraystretch}{1.3}
    \centering
    \caption{MNIST Training Metrics for Skeleton Fill(\%)}
    \begin{tabular}{| c | r | r | r | r | r |}
        \hline
        Class & \multicolumn{1}{|c|}{TP} & \multicolumn{1}{|c|}{TN} & \multicolumn{1}{|c|}{FP} & \multicolumn{1}{|c|}{FN} & \multicolumn{1}{|c|}{$ACC$} \\
        \hline
        \hline
        0 & 0.0 & 90.1 & 0.0 & 9.87 & 90.1 \\
        1 & 11.2 & 0.0 & 88.8 & 0.0 & 11.2 \\
        2 & 0.0 & 90.0 & 0.0 & 9.93 & 90.1 \\
        3 & 0.0 & 89.8 & 0.0 & 10.2 & 89.8 \\
        4 & 0.0 & 90.2 & 0.0 & 9.74 & 90.3 \\
        5 & 0.0 & 91.0 & 0.0 & 9.04 & 91.0 \\
        6 & 0.0 & 90.1 & 0.0 & 9.86 & 90.1 \\
        7 & 0.0 & 89.6 & 0.0 & 10.4 & 89.6 \\
        8 & 0.0 & 90.2 & 0.0 & 9.75 & 90.3 \\
        9 & 0.0 & 90.0 & 0.0 & 9.92 & 90.1 \\
        \hline
    \end{tabular}
    \label{tab:skel_fill_metrics}
\end{table}

\begin{table}
    \renewcommand{\arraystretch}{1.3}
    \centering
    \caption{Comparison of Various Effectiveness Metrics}
    \resizebox{\columnwidth}{!}{%
    \begin{tabular}{| l | c | c || c | c || c | c || c | c |}
        \cline{2-9}
        \multicolumn{1}{l}{} & \multicolumn{4}{| c ||}{MNIST} & \multicolumn{4}{ c |}{EMNIST} \\
        \cline{2-9}
        \multicolumn{1}{l}{} & \multicolumn{2}{| c ||}{MLP} & \multicolumn{2}{ c ||}{SVM} & \multicolumn{2}{ c ||}{MLP} & \multicolumn{2}{ c |}{SVM} \\
        \hline
        Effectiveness Metric & $E$ & $E+U$ & $E$ & $E+U$ & $E$ & $E+U$ & $E$ & $E+U$ \\
        \hline
        \hline
        $E_{PARS}$ & 95.5 & 97.6 & 95.4 & 97.3 & 71.7 & 77.4 & 75.9 & 81.0 \\
        $P \cdot R \cdot S$ & 95.2 & 97.4 & 95.1 & 97.1 & 71.7 & 77.3 & 75.9 & 81.0 \\
        $P \cdot R$ & 94.3 & 97.2 & 94.8 & 97.0 & 71.1 & 76.8 & 75.9 & 81.0 \\
        $P \cdot S$ & 95.0 & 97.3 & 94.4 & 96.8 & 70.4 & 76.9 & 74.7 & 80.8 \\
        Precision ($P$) & 94.4 & 97.0 & 94.2 & 96.7 & 70.4 & 76.8 & 74.7 & 80.8 \\
        Cohen's Kappa & 94.1 & 96.9 & 94.2 & 96.7 & 71.5 & 77.5 & 74.4 & 80.6 \\
        $MCC$ & 92.6 & 96.2 & 93.8 & 96.5 & 69.7 & 75.5 & 73.9 & 80.4 \\
        F1-Score & 91.9 & 95.9 & 93.5 & 96.4 & 70.6 & 76.7 & 74.2 & 80.5 \\
        $S \cdot R$ & 85.6 & 93.1 & 92.0 & 95.7 & 37.0 & 55.3 & 67.1 & 76.9 \\
        Specificity ($S$) & 88.1 & 93.7 & 92.0 & 95.1 & 49.2 & 61.5 & 68.5 & 76.9 \\
        Accuracy (ACC) & 85.6 & 92.6 & 92.0 & 95.1 & 47.7 & 60.0 & 68.0 & 76.3 \\
        AUC & 67.5 & 77.7 & 89.8 & 94.1 & 5.62 & 8.49 & 63.4 & 73.1 \\
        Balanced Accuracy & 75.0 & 85.4 & 89.7 & 94.2 & 8.46 & 16.4 & 63.1 & 72.9 \\
        Recall ($R$) & 52.4 & 68.0 & 85.1 & 91.7 & 2.53 & 3.35 & 50.5 & 63.7 \\
        \hline
    \end{tabular}
    }
    \label{tab:mnist_emnist_eff_metrics}
\end{table}

\begin{table*}
    \begin{minipage}{0.3\linewidth}
        \centering
        \includegraphics[width=2cm]{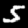}
        \captionof{figure}{A letter \lq{S}\rq~example.}
        \label{fig:ex4}
    \end{minipage}\hfill
    \begin{minipage}{0.7\linewidth}
        \renewcommand{\arraystretch}{1.3}
        \captionof{table}{Explainability Matrix based on the Recall Metric}\label{table:example4r}
        \centering
        \resizebox{\columnwidth}{!}{%
        \begin{tabular}{| c | c | c | c | c | c | c || c | c | c | c |}
        \cline{4-11}
        \multicolumn{3}{c}{} & \multicolumn{4}{|c||}{PDF Effectiveness} & \multicolumn{4}{c|}{PDF Explainability} \\
        \hline
        $F_j$ & Property & Class Vote & $E_{j,1}$ & $E_{j,E}$ & $E_{j,S}$ & $E_{j,n}$ & $Ex_1$ & $Ex_E$ & $Ex_S$ & $Ex_n$ \\
        \hline
        \hline
        $F_1$ & Stroke & \lq{\textit{S}}\rq &  &  & 0.9229 &  &  &  & 1.0 & \\ 
        \hline
        $F_2$ & Circle & \lq{\textit{1}}\rq & \bf{0.9954} &  &  &  & 1.0 &  &  & \\
        \hline
        $F_3$ & Crossing & \lq{\textit{1}}\rq & \bf{0.9850} &  &  &  & 1.0 &  &  & \\
        \hline
        $F_4$ & Ellipse & \lq{\textit{n}}\rq &  &  &  & 0.1079 &  &  &  & 1.0 \\
        \hline
        $F_5$ & Ell-Cir & \lq{\textit{1}}\rq & \bf{0.9954} &  &  &  &  &  & 1.0 & \\
        \hline
        $F_6$ & Endpoint & \lq{\textit{S}}\rq &  &  & 0.4963 &  &  &  & 1.0 & \\
        \hline
        $F_7$ & Encl. Reg. & \lq{\textit{1}}\rq & \bf{0.9996} &  &  &  & 1.0 &  &  & \\
        \hline
        $F_8$ & Line & \lq{\textit{S}}\rq &  &  & 0.4950 &  &  &  & 1.0 & \\
        \hline
        $F_9$& Convex Hull & \lq{\textit{E}}\rq &  & 0.2604 &  &  &  & 1.0 &  &  \\
        \hline
        $F_{10}$& Corner & \lq{\textit{S}}\rq &  &  & 0.5088 &  &  &  & 1.0 & \\
        \hline
        $F_{11}$& No Property & \lq{\textit{S}}\rq &  &  & 0.9238 &  &  &  & 0.0 & \\
        \hline
        \hline
        \multicolumn{3}{|c|}{Weights} & $3.975$ & $0.2604$ & $3.346$ & $0.1079$ & $3.975$ & $0.2604$  & $2.423$ & $0.1079$ \\
        \cline{0-10}
        \multicolumn{3}{|c|}{Confidence / Explainability} & $51.7\%$ & $3.39\%$ & $43.5\%$ & $1.40\%$ & $100\%$ & $100\%$ & $72.4\%$ & $100\%$ \\
        \cline{0-10}
        \end{tabular}
        }
    \end{minipage}

\end{table*}
    
\begin{table}[]
        \renewcommand{\arraystretch}{1.3}
        \captionof{table}{Explanation based on the Recall Metric} \label{table:exexample4r}
        \resizebox{\columnwidth}{!}{%
        \begin{tabular}{| m{0.05\columnwidth} | m{0.13\columnwidth} | m{0.16\columnwidth} | m{0.66\columnwidth} |}
        \hline
         Class & Confidence & Explainability & Explainable Description \\
        \hline \hline
        \lq{\textit{1}}\rq & 51.7\% & 100\% & Confidence is medium for interpreting this character as a one due to the enclosed region, circle, ellipse-circle, and crossing properties. \\ 
        \hline
        \lq{\textit{S}}\rq & 43.5\% & 72.4\% & Confidence is medium for interpreting this character as an S due to the stroke, corner, endpoint, and line properties. \\
        \hline
        \lq{\textit{E}}\rq & 3.39\% & 100\% & Confidence is low for interpreting this character as an E due to the convex hull property. \\
        \hline
        \lq{\textit{n}}\rq & 1.40\% & 100\% & Confidence is low for interpreting this character as an n due to the ellipse property. \\
        \hline
        \end{tabular}
        }
\end{table}

\section{Method} \label{method}

\subsection{Explainable Architecture} \label{explainable_arch}

Fig.~\ref{fig:xai_arch} depicts the explainable architecture with explainable
and unexplainable contributions.  Consider each horizontal element in the
architecture incident on the Decision Making Process as a Pre-Decision Flow
(PDF) of the explainable system.  An explainable PDF would consist of one
Property Transform and one Inference Engine (IE).  An unexplainable PDF would
consist of only one IE.  The $1$ through $n$ explainable PDFs in the
architecture are depicted as green while the single unexplainable PDF is
represented in orange as $IE_u$. Explainable PDFs are related to properties that
contribute to the explainability of the system. Often, explainable properties
provide mediocre recognition results compared to an IE trained against the
untransformed data.  In many cases, this is because a property may not pertain to
each class. E.g., the ellipse property does not apply to a well-formed digit
one.

In this architecture, the unexplainable PDF is the untransformed input image. The
unexplainable flow contributes excellent recognition performance but deters from
the explainability of the system since the IE is considered an opaque box,
without an explainable property for justification of a decision. An
explainability metric is introduced to quantify the
impact of unexplainable PDFs on decisions in Section \ref{exp_metric}.

In Fig.~\ref{fig:xai_arch}, input flows into the system at the left. The initial
stage of the architecture, for explainable PDFs, is the Transform phase. In the
Transform phase, the input image is transformed according to the explainable
property. Transforms emphasize particular explainable properties in the input
and are essential to explainability.

The second stage of the architecture is the Property Inferencing stage.  In
Property Inferencing, IEs are used to make decisions on local data. IEs in this
work are Multilayer Perceptrons (MLP) or Support Vector Machines (SVM).
Explainable IEs are trained on transformed training data. The unexplainable IE,
$IE_u$, is trained on untransformed input.  Training data and training results
are stored in a knowledgebase (KB) as depicted at the bottom of
Fig.~\ref{fig:xai_arch}.

Local decisions from Property Inferencing are passed to Decision Making, where a
global decision is made.  This global decision is merely an ordering of the
classes voted on based upon Confidence.  The Confidence, $C_d$, is shown in
\eqref{eq:confidence} where $W_d$, in \eqref{eq:weighted_effectiveness},
represents the weighted effectiveness that voted for a particular class of $d$
over the sum of weighted effectiveness for all classes $k$ voted for.
Effectiveness, $E_{j, d}$, indicates how well an IE is at recognizing a class
and is obtained from data stored in the KB. Section \ref{effectiveness_metrics}
details effectiveness and outlines new methods of gauging effectiveness while
Section \ref{exp_results} provides results of the new effectiveness metrics.

\begin{equation}\label{eq:confidence}
    C_d = \frac{W_d}{\sum\limits_kW_k}
\end{equation}

\begin{equation}\label{eq:weighted_effectiveness}
    W_d=\sum_j E_{j, d}
\end{equation}

The Explainability phase in the explainable Architecture,
Fig.~\ref{fig:xai_arch}, composes the textual rationale, related to the
explainable properties contributing to decisions.  This is done by constructing
phrases indicating the properties that voted for each class.  The explainability
measure, $Ex_d$ from \eqref{eq:explainability}, is also determined in this phase
of the system. The measure indicates how explainable each decision for a class,
$d$, is based on the flow's relation to an explainable property.  The user is
finally presented with the ranked recognition results, the confidence of the
decisions, the rationale for a decision, and the level of explainability for
each decision. Sample explainable results are presented in Section
\ref{explainable_examples}.

The skeleton fill flow was not used due to poor performance, almost no $TP$s for
nine out of ten MNIST classes, as observed in Table~\ref{tab:skel_fill_metrics}. A new
property representing corners or inflection points in a character was
introduced.  The corner transform utilizes the Harris corner detection algorithm
\cite{harris1988combined}.

\subsection{Explainable metric for Unexplainable Flows} \label{exp_metric}

Explainability, in the property-based system, is provided by explainable
properties, used to justify a local decision in plain terms to a
user.  With the introduction of unexplainable flows, each flow, $j$, is assigned
an explainability metric $X_j$, where $ 0 \leq X_j \leq 1$.  An $X_j$ close to
zero would signify an unexplainable flow while an $X_j$ near one would indicate an
explainable flow.  Examples in this work use $X_d=1$ for explainable and
$X_d=0$ for the unexplainable flows.

\begin{equation}\label{eq:explainability}
 Ex_d=\frac{\sum{E_{j,d}X_j}}{\sum{E_{j,d}}}
\end{equation}

The explainability of a decision for class $d$ is given by $Ex_d$ in
\eqref{eq:explainability} where the numerator is the sum of the product of the
explainability metric and effectiveness for the flows that voted for class $d$, and
the denominator is the sum of effectiveness for flows that voted for class $d$.

\subsection{Effectiveness Metrics} \label{effectiveness_metrics}

\begin{table*}[]
    \renewcommand{\arraystretch}{1.3}
    \captionof{table}{Explainability Matrix based on the Accuracy Metric}\label{table:example4a}
    \centering
    \begin{tabular}{| c | c | c | c | c | c | c || c | c | c | c |}
    \cline{4-11}
    \multicolumn{3}{c}{} & \multicolumn{4}{|c||}{PDF Effectiveness} & \multicolumn{4}{c|}{PDF Explainability} \\
    \hline
    $F_j$ & Property & Class Vote & $E_{j,1}$ & $E_{j,E}$ & $E_{j,S}$ & $E_{j,n}$ & $Ex_1$ & $Ex_E$ & $Ex_S$ & $Ex_n$ \\
    \hline
    \hline
    $F_1$ & Stroke & \lq{\textit{S}}\rq &  &  & 0.9964 &  &  &  & 1.0 & \\ 
    \hline
    $F_2$ & Circle & \lq{\textit{1}}\rq & 0.8417 &  &  &  & 1.0 &  &  & \\
    \hline
    $F_3$ & Crossing & \lq{\textit{1}}\rq &  0.7354 &  &  &  & 1.0 &  &  & \\
    \hline
    $F_4$ & Ellipse & \lq{\textit{n}}\rq &  &  &  & 0.9774 &  &  &  & 1.0 \\
    \hline
    $F_5$ & Ell-Cir & \lq{\textit{1}}\rq & 0.8444 &  &  &  &  &  & 1.0 & \\
    \hline
    $F_6$ & Endpoint & \lq{\textit{S}}\rq &  &  & 0.9776 &  &  &  & 1.0 & \\
    \hline
    $F_7$ & Encl. Reg. & \lq{\textit{1}}\rq & 0.3615 &  &  &  & 1.0 &  &  & \\
    \hline
    $F_8$ & Line & \lq{\textit{S}}\rq &  &  & 0.9787 &  &  &  & 1.0 & \\
    \hline
    $F_9$& Convex Hull & \lq{\textit{E}}\rq &  & 0.9767 &  &  &  & 1.0 &  &  \\
    \hline
    $F_{10}$& Corner & \lq{\textit{S}}\rq &  &  & 0.9791 &  &  &  & 1.0 & \\
    \hline
    $F_{11}$& No Property & \lq{\textit{S}}\rq &  &  & 0.9964 &  &  &  & 0.0 & \\
    \hline
    \hline
    \multicolumn{3}{|c|}{Weights} & $2.783$ & $0.9767$ & $4.928$ & $0.9774$ & $2.783$ & $0.9767$  & $3.932$ & $0.9774$ \\
    \cline{0-10}
    \multicolumn{3}{|c|}{Confidence / Explainability} & $28.8\%$ & $10.1\%$ & $51.0\%$ & $10.1\%$ & $100\%$ & $100\%$ & $79.8\%$ & $100\%$ \\
    \cline{0-10}
    \end{tabular}
\end{table*}

\begin{table}[htbp]
    \renewcommand{\arraystretch}{1.3}
    \captionof{table}{Explanations based on the Accuracy Metric}\label{table:exexample4a}
    \resizebox{\columnwidth}{!}{%
    \begin{tabular}{| m{0.05\columnwidth} | m{0.13\columnwidth} | m{0.16\columnwidth} | m{0.66\columnwidth} |}
    \hline
     Class & Confidence & Explainability & Explainable Description \\
    \hline \hline
    \lq{\textit{S}}\rq & 51.0\% & 79.8\% & Confidence is medium for interpreting this character as an S due to the stroke, corner, endpoint, and line properties. \\ 
    \hline
    \lq{\textit{1}}\rq & 28.8\% & 100\% & Confidence is medium for interpreting this character as a one due to the enclosed region, circle, ellipse-circle, and crossing properties. \\
    \hline
    \lq{\textit{n}}\rq & 10.1\% & 100\% & Confidence is low for interpreting this character as an n due to the ellipse property. \\
    \hline
    \lq{\textit{E}}\rq & 10.1\% & 100\% & Confidence is low for interpreting this character as an E due to the convex hull property. \\
    \hline
    \end{tabular}
    }
\end{table}

An important area of the architecture, influencing the overall performance,
concerns assigning weights to the IE votes. A key concept, related to assigning
of weights to the IE votes, is the Effectiveness of an IE at recognizing a
particular class. Effectiveness, $E_{j,d}$, was posed as a metric for how well
the $j^{th}$ IE performs at recognizing class $d$ and is used to weight IE
votes. Recall, given in \eqref{eq:recall},  was previously used for
Effectiveness. While Recall had acceptable results, the system exhibited false
positives for some of the poor-performing flows.





It is not enough to know that an IE is effective overall. Effectiveness must be
examined at the class level. This requires that metrics are taken as one class
versus others. The one versus others strategy imposes a significant imbalance
between the single class and others with the MNIST and EMNIST datasets
\cite{deng2012mnist, cohen2017emnist}.  Some metrics are sensitive to imbalanced
data and produce misleading results. This is evident in
Table~\ref{tab:skel_fill_metrics} where Accuracy ($ACC$), for nine out of ten
classes, is very high, around 90\% despite no $TP$.


Particular metrics from literature that were examined and evaluated as
effectiveness in the explainable architecture include Accuracy, Recall,
Specificity, Precision, F1-Score, Cohen's Kappa, Matthews Correlation
Coefficient, Balanced Accuracy, and the Area Under the Curve of the Receiver
Operator Characteristic (AUC).


In addition to evaluating metrics from literature, some metrics were combined
and compared.  The combination of metrics found to have the best performance as
per-class effectiveness was the product of class vs others Precision ($P$),
Accuracy ($ACC$), Recall ($R$), and Specificity ($S$) as shown in
\eqref{eq:epars} as the new effectiveness metric, $E_{PARS}$.
Equation~\eqref{eq:epars_expansion} shows the expansion of $E_{PARS}$ in terms of
$TP$, $TN$, $FP$, and $FN$.

\begin{equation}\label{eq:epars}
    E_{PARS} = P \cdot ACC \cdot R \cdot S
\end{equation}

\small
\begin{equation}\label{eq:epars_expansion}
\frac{TN {\cdot} TP^3+TN^2 {\cdot} TP^2}{(TN{+}FP)(TP{+}FP)(TP{+}FN)(TP{+}TN{+}FP{+}FN)}
\end{equation}
\normalsize

\subsection{Inference Engine Training} \label{svm}

SVM IEs used the Support Vector Classification implementation and NN IEs used
the MLP Classifier implementation in scikit-learn version 1.2.2
\cite{scikit-learn}.  In the SVMs, radial basis function kernels were used as
they performed better than alternative kernels with the transformed data.
MLPs used two hidden layers of 128 neurons with a rectified linear unit
activation function.  This MLP architecture performed well on the
handwritten data and was fast to train without specialized hardware.



Results shown in examples were obtained using discrete classification of the MLP
and SVM models, which outputs either a one or zero, not continuous probability
estimates p,  where $0 \leq p \leq 1$.  The default behavior for the
scikit-learn SVM model is discrete classification.  When the SVMs were trained to
enable probability estimates and the estimates were used along with
effectiveness, there was much less variance between the metrics and higher
system accuracy.

\section{Results} \label{results}

Table~\ref{tab:mnist_emnist_eff_metrics} depicts the MNIST and EMNIST overall
accuracy percentage results obtained with MLP and SVM using various per-class
effectiveness metrics in ranking and selecting a global decision. MLPs and SVMs
performed comparably on MNIST and the SVMs performed a few percentage points
better on EMNIST.  SVMs also appeared to be more forgiving when used with lower
performing effectiveness metrics. 

Accuracy reflected in the tables is an appropriate overall measure of
performance of the system since the classes are balanced and the metric is being
taken on the entire architecture. The Explainable result (E) columns indicate
the overall system accuracy for ten explainable flows while the combined
Explainable and Unexplainable result (E+U) columns indicate the overall
accuracy using ten explainable and one unexplainable flow.

\subsection{Explainable Results} \label{exp_results}

Adding an accurate, but unexplainable, classifier to the system improves
performance. This is illustrated in Table \ref{tab:mnist_emnist_eff_metrics}
where combined results (E+U) are greater than the strictly explainable ($E$)
results. A marked improvement in accuracy was shown by moving to combined
explainable and unexplainable flows.

The combined (E+U) results were greater than explainable (E) in all cases.
In the MNIST example using Recall as Effectiveness, the increase using MLP flows
was over 15\%.  Typical increases with combined flows ranged from two to ten
percentage points better than strictly explainable flows.


\subsection{Effectiveness Results} \label{eff_results}

\begin{table*}[]
    \renewcommand{\arraystretch}{1.3}
    \captionof{table}{Explainability Matrix based on the $E_{PARS}$ Metric}\label{table:example4epars}
    \centering
    \begin{tabular}{| c | c | c | c | c | c | c || c | c | c | c |}
    \cline{4-11}
    \multicolumn{3}{c}{} & \multicolumn{4}{|c||}{PDF Effectiveness} & \multicolumn{4}{c|}{PDF Explainability} \\
    \hline
    $F_j$ & Property & Class Vote & $E_{j,1}$ & $E_{j,E}$ & $E_{j,S}$ & $E_{j,n}$ & $Ex_1$ & $Ex_E$ & $Ex_S$ & $Ex_n$ \\
    \hline
    \hline
    $F_1$ & Stroke & \lq{\textit{S}}\rq &  &  & 0.8341 &  &  &  & 1.0 & \\ 
    \hline
    $F_2$ & Circle & \lq{\textit{1}}\rq & 0.0830 &  &  &  & 1.0 &  &  & \\
    \hline
    $F_3$ & Crossing & \lq{\textit{1}}\rq & 0.0389 &  &  &  & 1.0 &  &  & \\
    \hline
    $F_4$ & Ellipse & \lq{\textit{n}}\rq &  &  &  & 0.0406 &  &  &  & 1.0 \\
    \hline
    $F_5$ & Ell-Cir & \lq{\textit{1}}\rq & 0.0848 &  &  &  &  &  & 1.0 & \\
    \hline
    $F_6$ & Endpoint & \lq{\textit{S}}\rq &  &  & 0.2274 &  &  &  & 1.0 & \\
    \hline
    $F_7$ & Encl. Reg. & \lq{\textit{1}}\rq & 0.0040 &  &  &  & 1.0 &  &  & \\
    \hline
    $F_8$ & Line & \lq{\textit{S}}\rq &  &  & 0.2401 &  &  &  & 1.0 & \\
    \hline
    $F_9$& Convex Hull & \lq{\textit{E}}\rq &  & 0.1067 &  &  &  & 1.0 &  &  \\
    \hline
    $F_{10}$& Corner & \lq{\textit{S}}\rq &  &  & 0.2508 &  &  &  & 1.0 & \\
    \hline
    $F_{11}$& No Property & \lq{\textit{S}}\rq &  &  & 0.8385 &  &  &  & 0.0 & \\
    \hline
    \hline
    \multicolumn{3}{|c|}{Weights} & $0.2106$ & $0.1067$ & $2.391$ & $0.0406$ & $0.2106$ & $0.1067$  & $1.5524$ & $0.0406$ \\
    \cline{0-10}
    \multicolumn{3}{|c|}{Confidence  / Explainability} & $7.66\%$ & $3.88\%$ & \bf{86.9\%} & $1.48\%$ & $100\%$ & $100\%$ & \bf{64.9\%} & $100\%$ \\
    \cline{0-10}
    \end{tabular}
\end{table*}

\begin{table}[]
    \renewcommand{\arraystretch}{1.3}
    \captionof{table}{Explanations based on the $E_{PARS}$ Metric}\label{table:exexample4epars}
    \resizebox{\columnwidth}{!}{%
    \begin{tabular}{| m{0.05\columnwidth} | m{0.13\columnwidth} | m{0.16\columnwidth} | m{0.66\columnwidth} |}
    \hline
     Class & Confidence & Explainability & Explainable Description \\
    \hline \hline
    \lq{\textit{S}}\rq & 86.9\% & 64.9\% & Confidence is high for interpreting this character as an S due to the stroke, corner, endpoint, and line properties. \\ 
    \hline
    \lq{\textit{1}}\rq & 7.66\% & 100\% & Confidence is low for interpreting this character as a one due to the enclosed region, circle, ellipse-circle, and crossing properties. \\
    \hline
    \lq{\textit{E}}\rq & 3.88\% & 100\% & Confidence is low for interpreting this character as an E due to the convex hull property. \\
    \hline
    \lq{\textit{n}}\rq & 1.40\% & 100\% & Confidence is low for interpreting this character as an n due to the ellipse property. \\
    \hline
    \end{tabular}
    }
\end{table}   

Some of the performance metrics from the literature, especially those that are
resilient to imbalanced data such as Cohen's Kappa, provide outstanding results
as a measure of effectiveness in the explainable architecture. Surprisingly,
Precision alone as well as the product of Precision and other metrics perform
among the highest of those attempted. The best-performing metric found is given
in \eqref{eq:epars} as $E_{PARS}$, the product of Precision, Accuracy, Recall,
and Specificity.  The previously reported accuracy observed, using the
explainable architecture, on MNIST with MLPs was about 92\% using Recall as
Effectiveness and probability estimates. Employing $E_{PARS}$ as effectiveness
and without using probability estimates, accuracy was observed at about 95.5\%,
an improvement of over 3\%. When probability estimates are used with $E_{PARS}$
as effectiveness, accuracy was increased to 96.1\%.  Adding an unexplainable
component to the system and using probability estimates increased accuracy
results to as high as 98.0\%. 

Observing the numerator in \eqref{eq:epars_expansion}, $E_{PARS}$
performs well because $TP$ to $TN$ balance is maintained since $TP^3$
($\approx2.2\text{x}10^{11}$ for MNIST and $\approx1.4\text{x}10^{10}$ for
EMNIST) and $TN^2$ ($\approx2.7\text{x}10^9$ for MNIST and
$\approx1.2\text{x}10^{10}$ for EMNIST).  A larger dataset with more classes may
not result in similar performance.

\subsection{Explainable Example} \label{explainable_examples}

This section presents an example and outlines how the explainable architecture
comes to a decision.  The example illustrates the confidence, rationale, and
explainability provided to a user.  Also included are results with differing
effectiveness metrics.  Where the terms low, medium, and high are used in this
section, they denote below 25\%, 25\% to 75\%, and above 75\%, respectively.

Fig.~\ref{fig:ex4} depicts the example, an EMNIST balanced test, handwritten
letter \lq{S}\rq~sample (index 18) labeled as a capital \lq{S}\rq.  The explainable
architecture used for the example is composed of eleven PDFs with SVM IEs. Data
from multiple effectiveness metrics (such as Recall, Accuracy, and $E_{PARS}$)
will be presented to demonstrate how the explainable system benefits from more
robust effectiveness metrics.

The results of processing the example using the Recall metric are in Tables
\ref{table:example4r} and \ref{table:exexample4r}, the results of using the
Accuracy metric are depicted in Tables \ref{table:example4a} and
\ref{table:exexample4a}, and the results of using the $E_{PARS}$ metric are in
Tables \ref{table:example4epars} and \ref{table:exexample4epars}.  Table
\ref{table:example4eff_metrics} illustrates the metric particulars for the PDFs
in the example in one place.  Note that four classes were voted on for the
example.  Each PDF's vote and explainability metric is the same across the
various tables. Only the effectiveness varies.

When examining the Explainability data from Tables \ref{table:example4r},
\ref{table:example4a}, and \ref{table:example4epars}, each flow identifier,
$F_j$, is in the first column where $j$ denotes the flow number.  The second
column indicates the property name. Flow $F_{11}$ is labeled as no property
because it represents an unexplainable flow, without an explainable property.
The Class Vote column indicates the class that was selected by each flow.

The remaining columns are related to effectiveness and explainability.  Since
four classes were voted on in this example, there will be four columns each for
effectiveness and explainability, representing each class with a vote. The
columns labeled $E_{j,d}$, represent the effectiveness of the flows for each
class, $d$, that were voted upon. The columns labeled $Ex_d$ represent the
explainability metric for each flow.  Only the column representing the class
that a flow voted for will have a value in the table for effectiveness and
explainability.  The last two rows of the Explainability tables represent the
weights, of effectiveness and explainability, as well as the Confidence, $C_d$,
and Explainability, $Ex_d$, for each class $d$.

Ordered explanation results are represented in Tables \ref{table:exexample4r},
\ref{table:exexample4a}, and \ref{table:exexample4epars}.  The first column
indicates the class, $d$. The second column the Confidence.  The third column,
represents the explainability associated with the decision. The final column is
the rationale provided to the user.

Observe in Tables \ref{table:example4r} and \ref{table:exexample4r} that the
four PDFs that voted for class one had high Recall, reflected in the
Effectiveness column $E_{j,1}$ of Table \ref{table:example4r} in bold.  This
resulted in the digit one winning with medium confidence, $51.7\%$, due to the
enclosed region, circle, ellipse-circle, and crossing properties.  The second
choice was the \lq{S}\rq~with medium confidence, $43.5\%$. Since Recall in
\eqref{eq:recall} has $TP$ in the numerator, high $TP$ and very low $FN$ counts of the
PDFs that voted for the digit one observed in Table
\ref{table:example4eff_metrics} explains why the digit one wins using the Recall
metric.



Tables \ref{table:example4a} and \ref{table:exexample4a} contain data and
results from using the Accuracy metric on the example.  Accuracy is given in
\eqref{eq:accuracy}.  Due to the comparatively high $TN$ rate of the PDFs that
voted for the \lq{S}\rq, as shown in bold in Table
\ref{table:example4eff_metrics} it wins with medium confidence, $51.0\%$, due to
the stroke, corner, endpoint, and line properties. The \lq{S}\rq~also received a
vote from the unexplainable PDF.  The lower explainability metric, at $79.8\%$,
reflects diminished explainability due to no explainable property associated
with part ($20.2\%$) of the contribution for the decision.


\begin{table}[htbp]
    \renewcommand{\arraystretch}{1.3}
    \captionof{table}{Ex.~2 Effectiveness Metrics (\%) based on Votes}\label{table:example4eff_metrics} 
    \centering
    \resizebox{\columnwidth}{!}{%
    \begin{tabular}{| c | c | c | c | c | c | c | c | c | c | c | c | c | c |}
    \hline
    Property & Class & TP & TN & FP & FN & R & ACC & $E_{PARS}$ \\
    \hline
    \hline
    Encl. Reg. & \lq{\textit{1}}\rq & 2.13 & 34.0 & 63.8 & 0.0 & 100 & 36.2 & 0.40 \\ 
    \hline
    Circle & \lq{\textit{1}}\rq & 2.12 & 82.1 & 15.8 & 0.0 & 99.5 & 84.2 &  8.29 \\
    \hline
    Ell-Cir & \lq{\textit{1}}\rq & 2.12 & 82.3 & 15.5 & 0.0 & 99.5 & 84.5 & 8.47 \\
    \hline
    Crossing & \lq{\textit{1}}\rq & 2.01 & 71.5 & 26.4 & 0.0 & 98.5 & 73.5 & 3.88 \\
    \hline
    Convex Hull & \lq{\textit{E}}\rq & 0.55 & 97.1 & 0.76 & 1.57 & 26.0 & 97.7 & 10.7 \\
    \hline
    No Property & \lq{\textit{S}}\rq & 1.97 & \bf{97.7} & 0.19 & 0.16 & 92.4 & 99.7 & 83.9 \\
    \hline
    Stroke & \lq{\textit{S}}\rq & 1.96 & \bf{97.7} & 0.20 & 0.16 & 92.3 & 99.6 & 83.4 \\
    \hline
    Corner & \lq{\textit{S}}\rq & 1.08 & \bf{96.8} & 1.04 & 1.05 & 50.9 & 97.9 & 25.1 \\
    \hline
    Endpoint & \lq{\textit{S}}\rq & 1.06 & \bf{96.7} & 0.20 & 0.20 & 49.6 & 97.8 & 22.7 \\
    \hline
    Line & \lq{\textit{S}}\rq & 1.05 & 96.8 & 1.05 & 1.07 & 49.5 & 97.9 & 24.0 \\
    \hline
    Ellipse & \lq{\textit{n}}\rq & 0.23 & 97.5 & 0.36 & 1.90 & 10.8 & 97.7 & 4.07 \\
    \hline
    \end{tabular}
    }
\end{table}

The final set of Explainability and Explanations Tables,
\ref{table:example4epars} and \ref{table:exexample4epars}, involve the
$E_{PARS}$ metric. The stroke and unexplainable flows have a much higher
effectiveness than other metrics.  High effectiveness from the Accuracy
metric for the flows voting for the digit one were due to insensitivity of
Accuracy to $FP$s.  The same flows voting for the digit one have a reduced
effectiveness due to relative sensitivity of $E_{PARS}$ to $FP$ counts, as
observed in Table \ref{table:example4eff_metrics}.  The capital \lq{S}\rq~
correctly wins with high confidence, $86.9\%$, using $E_{PARS}$.

\section*{Conclusion}

Introduction of unexplainable, but high-performing, flows into the explainable
architecture increased the accuracy and explainability of the system.  Using
previous effectiveness metrics an increase in MNIST accuracy with MLP was
observed at over 15\% by introducing unexplainable flows.  A metric to
characterize the impact of the unexplainable addition to the system, $Ex_d$, and
along with an example from EMNIST, illustrate its utility.

The results and analysis pertaining to performance metrics to gauge
effectiveness suggest that several metrics from the literature perform better
than previous effectiveness in the explainable architecture, as noted in Section
\ref{eff_results}. The best-performing metric $E_{PARS}$ was devised as the
product of Precision, Accuracy, Recall, and Specificity.  The example
demonstrated that more robust and resilient effectiveness metrics improve
results.  With analysis of the $E_{PARS}$ metric, one could speculate that
larger data sets, especially with more classes, may not perform as well with
$E_{PARS}$. A derivation of $E_{PARS}$ that similarly scales the $TP$ at or above
$TN$ based on the dataset size and number of classes could be devised and may be
suitable as a metric gauging model performance on imbalanced datasets.



\bibliographystyle{IEEEtran}
\bibliography{references}

\begin{thebibliography}{10}
\providecommand{\url}[1]{#1}
\csname url@samestyle\endcsname
\providecommand{\newblock}{\relax}
\providecommand{\bibinfo}[2]{#2}
\providecommand{\BIBentrySTDinterwordspacing}{\spaceskip=0pt\relax}
\providecommand{\BIBentryALTinterwordstretchfactor}{4}
\providecommand{\BIBentryALTinterwordspacing}{\spaceskip=\fontdimen2\font plus
\BIBentryALTinterwordstretchfactor\fontdimen3\font minus
  \fontdimen4\font\relax}
\providecommand{\BIBforeignlanguage}[2]{{%
\expandafter\ifx\csname l@#1\endcsname\relax
\typeout{** WARNING: IEEEtran.bst: No hyphenation pattern has been}%
\typeout{** loaded for the language `#1'. Using the pattern for}%
\typeout{** the default language instead.}%
\else
\language=\csname l@#1\endcsname
\fi
#2}}
\providecommand{\BIBdecl}{\relax}
\BIBdecl

\bibitem{deng2012mnist}
L.~Deng, ``The mnist database of handwritten digit images for machine learning
  research,'' \emph{IEEE Signal Processing Magazine}, vol.~29, no.~6, pp.
  141--142, 2012.

\bibitem{cohen2017emnist}
G.~Cohen, S.~Afshar, J.~Tapson, and A.~Van~Schaik, ``Emnist: Extending mnist to
  handwritten letters,'' in \emph{2017 international joint conference on neural
  networks (IJCNN)}.\hskip 1em plus 0.5em minus 0.4em\relax IEEE, 2017, pp.
  2921--2926.

\bibitem{6797059}
R.~A. Jacobs, M.~I. Jordan, S.~J. Nowlan, and G.~E. Hinton, ``Adaptive mixtures
  of local experts,'' \emph{Neural Computation}, vol.~3, no.~1, pp. 79--87,
  1991.

\bibitem{perrone1993putting}
M.~Perrone, ``Putting it all together: Methods for combining neural networks,''
  \emph{Advances in neural information processing systems}, vol.~6, 1993.

\bibitem{sharkey1996combining}
A.~J.~C. SHARKEY, ``On combining artificial neural nets,'' \emph{Connection
  science}, vol.~8, no. 3-4, pp. 299--314, 1996.

\bibitem{vaughan2018explainable}
J.~Vaughan, A.~Sudjianto, E.~Brahimi, J.~Chen, and V.~N. Nair, ``Explainable
  neural networks based on additive index models,'' \emph{arXiv preprint
  arXiv:1806.01933}, 2018.

\bibitem{whitten21}
P.~Whitten, F.~Wolff, and C.~n. Papachristou, ``Explainable artificial
  intelligence methodology for handwritten applications,'' in \emph{NAECON 2021
  - IEEE National Aerospace and Electronics Conference}, 2021, pp. 277--282.

\bibitem{whitten23}
P.~Whitten, F.~Wolff, and C.~Papachristou, ``Explainable neural network
  recognition of handwritten characters,'' in \emph{2023 IEEE 13th Annual
  Computing and Communication Workshop and Conference (CCWC)}, 2023, pp.
  0176--0182.

\bibitem{harter1971cranfield}
S.~P. Harter, ``The cranfield ii relevance assessments: A critical
  evaluation,'' \emph{The Library Quarterly}, vol.~41, no.~3, pp. 229--243,
  1971.

\bibitem{picek2019curse}
S.~Picek, A.~Heuser, A.~Jovic, S.~Bhasin, and F.~Regazzoni, ``The curse of
  class imbalance and conflicting metrics with machine learning for
  side-channel evaluations,'' \emph{IACR Transactions on Cryptographic Hardware
  and Embedded Systems}, pp. 209--237, 2019.

\bibitem{erickson2021magician}
B.~J. Erickson and F.~Kitamura, ``Magician’s corner: 9. performance metrics
  for machine learning models,'' p. e200126, 2021.

\bibitem{Naser_2021}
M.~Z. Naser and A.~H. Alavi, ``Error metrics and performance fitness indicators
  for artificial intelligence and machine learning in engineering and
  sciences,'' \emph{Architecture, Structures and Construction}, vol.~3, no.~4,
  p. 499–517, Nov. 2021.

\bibitem{METZ1978283}
C.~E. Metz, ``Basic principles of roc analysis,'' \emph{Seminars in Nuclear
  Medicine}, vol.~8, no.~4, pp. 283--298, 1978.

\bibitem{Hanley_1982}
J.~A. Hanley and B.~J. McNeil, ``The meaning and use of the area under a
  receiver operating characteristic (roc) curve.'' \emph{Radiology}, vol. 143,
  no.~1, pp. 29--36, 1982.

\bibitem{sasaki2007truth}
Y.~Sasaki \emph{et~al.}, ``The truth of the f-measure,'' \emph{Teach tutor
  mater}, vol.~1, no.~5, pp. 1--5, 2007.

\bibitem{Cohen1960ACO}
J.~Cohen, ``A coefficient of agreement for nominal scales,'' \emph{Educational
  and Psychological Measurement}, vol.~20, pp. 37 -- 46, 1960.

\bibitem{ben2008relationship}
A.~Ben-David, ``About the relationship between roc curves and cohen's kappa,''
  \emph{Engineering Applications of Artificial Intelligence}, vol.~21, no.~6,
  pp. 874--882, 2008.

\bibitem{MATTHEWS1975442}
B.~Matthews, ``Comparison of the predicted and observed secondary structure of
  t4 phage lysozyme,'' \emph{Biochimica et Biophysica Acta (BBA) - Protein
  Structure}, vol. 405, no.~2, pp. 442--451, 1975.

\bibitem{9385097}
D.~Chicco, V.~Starovoitov, and G.~Jurman, ``The benefits of the matthews
  correlation coefficient (mcc) over the diagnostic odds ratio (dor) in binary
  classification assessment,'' \emph{IEEE Access}, vol.~9, pp.
  47\,112--47\,124, 2021.

\bibitem{harris1988combined}
C.~Harris, M.~Stephens \emph{et~al.}, ``A combined corner and edge detector,''
  in \emph{Alvey vision conference}, vol.~15, no.~50.\hskip 1em plus 0.5em
  minus 0.4em\relax Citeseer, 1988, pp. 10--5244.

\bibitem{scikit-learn}
F.~Pedregosa, G.~Varoquaux, A.~Gramfort, V.~Michel, B.~Thirion, O.~Grisel,
  M.~Blondel, P.~Prettenhofer, R.~Weiss, V.~Dubourg, J.~Vanderplas, A.~Passos,
  D.~Cournapeau, M.~Brucher, M.~Perrot, and E.~Duchesnay, ``Scikit-learn:
  Machine learning in {P}ython,'' \emph{Journal of Machine Learning Research},
  vol.~12, pp. 2825--2830, 2011.

\end{thebibliography}

\end{document}